\documentclass[sigconf]{acmart}

\newcommand{\srl}{\textsc{srl}}
\newcommand{\cond}{\textsc{io}}
\newcommand{\sd}[1]{\scriptsize{$\pm$#1}}

\usepackage{booktabs}
\usepackage{multirow}
\usepackage{subcaption}

\usepackage{algorithm}
\usepackage{algorithmic}

\AtBeginDocument{
  }

\setcopyright{acmlicensed}
\copyrightyear{2026}
\acmYear{2026}
\acmDOI{}
\acmConference[ICAIL 2026]{the International Conference on Artificial Intelligence and Law}{June 08--12, 2026}{Singapore}
\acmISBN{}
\begin{document}

\title[From Legal Text to Executable Decision Models]{From Legal Text to Executable Decision Models: Evaluating Structured Representations for Legal Decision Model Generation}

\author{David Graus}
\orcid{0000-0002-6245-0870}
\affiliation{
   \institution{University of Amsterdam}
   \city{Amsterdam}
   \country{The Netherlands}
}
\email{d.p.graus@uva.nl}

\begin{abstract}
Transforming legal text into executable decision logic is a longstanding challenge in legal informatics.
With the rise of LLMs, this task has gained renewed interest, but remains challenging due to requiring extensive manual coding and evaluation.
In this paper, we employ a unique real-world dataset that pairs production-grade decision models with legal text from the Dutch Environment and Planning Act.
These models power the Omgevingsloket, a government platform where citizens check permit requirements for environmental activities.
We leverage this dataset to study whether intermediate structured representations can improve LLM-based generation of executable decision models from legal text.
We compare LLM generation under four input conditions: raw legal text, text enriched with semantic role labels, text enriched with input and output constraints, and legal text enriched with both.
We evaluate generated decision models along two dimensions:
\emph{structural evaluation}, through similarity to gold decision models with graph kernels and graphs' descriptive statistics, and
\emph{outcome evaluation} through functional equivalence by executing models on pre-configured test scenarios.
Our findings show that I/O constraints provide the dominant improvement (+37-54\% similarity over baseline), while semantic role labels alone show modest improvement depending on decision model type.
Outcome evaluation shows that on average, generated models match the gold
standard on 51-53\% of test scenarios, even though the generated models are typically smaller with lower decision complexity.
Our analysis finds LLMs can simplify decision models by eliminating redundant pass-through logic that comprises up to 45–55\% of nodes.
Importantly, structural similarity and outcome equivalence are complementary: high structural similarity does not guarantee outcome equivalence, and vice versa.
Our findings suggest that LLM-generated models can reduce manual effort in maintaining regulatory decision systems when legal experts provide interface specifications, though human verification remains essential.
To facilitate reproducibility, we publicly release our dataset of 95
production decision models with associated legal text and all experimental
code.
\end{abstract}


\maketitle

\section{Introduction}
\label{sec:intro}

Translating legal text into machine-readable and executable decision models is a longstanding challenge in legal informatics~\cite{sergot1986british}; highly contextual and structurally implicit language makes direct translation difficult, even for human experts.
While recent advances in large language models (LLMs) offer new opportunities, existing work still typically relies on manual coding and hand-curated rule sets~\cite{zin2025machine,11190345}.

An open question is how to help LLMs to reliably generate decision models that model regulatory logic.
One approach is to enrich the \emph{source}, e.g., annotating legal text with semantic role labels to aid interpretation.
Another approach, known from LLM-powered code generation, is to specify the \emph{target} by defining the interface (expected inputs and/or desired outputs).
These represent two different strategies: helping the model understand the source, or steering what to generate.

In this paper, we study both source enrichment (with semantic role labels) and target specification (through interface constraints) for LLM-based legal model generation.
We use a unique dataset from the Dutch Environment and Planning Act\footnote{\url{https://www.government.nl/topics/environment-and-planning-act}} (\emph{Omgevingswet}) that pairs legal text with hand-crafted executable decision models.
The decision models power the act's associated Environment and Planning Portal (\emph{Omgevingsloket}), a government platform where citizens can check requirements, apply for permits, or submit notifications of their environmental activities.
We discuss the nature of the models and act in Section~\ref{sec:method}.

We compare four different input representations for LLM-powered decision model generation from legal text:
\begin{itemize}
    \item \textbf{Text}: Raw legal article text only
    \item \textbf{Text+\srl{}}: Legal article text, source-enriched with semantic role labels (actors, actions, objects, recipients)
    \item \textbf{Text+\cond{}}: Legal article text, target-constrained with I/O (input and output) specifications derived from the decision model
    \item \textbf{Text+\srl{}+\cond{}}: Combined semantic role labels and I/O specifications
\end{itemize}

We evaluate generated decision models on
(i) their structural similarity, and
(ii) outcome equivalence (through model execution)
to gold models.
We evaluate with real-world, production-grade gold standard decision models, used by citizens and businesses to comply with environmental regulations.
This real-world scenario provides validity that is often absent in automated legal reasoning research, but also highlights practical challenges: the models encode actual regulatory complexity rather than idealized representations, and our findings directly inform systems that impact legal compliance in practice.

We make the following contributions:
(i) we present a controlled evaluation of structured representations for law-to-code generation using 95 real-world legal decision models,
(ii) we find that I/O specifications substantially improve generation quality (+37--54\% structural similarity), while semantic role labels provide minimal additional benefit,
(iii) we analyze the relationship between structural similarity and outcome equivalence and find they are complementary,
(iv) we find that LLMs validly simplify models by eliminating redundant modeling conventions,
(v) we publicly release our dataset and code to facilitate reproducibility.\footnote{\url{https://github.com/opengov-lab/legal-text-to-decision-model}}

\section{Related Work}
\label{sec:related}

\subsection{Legal Knowledge Representation}
Legal knowledge has been formalized using various representations, with tradeoffs between expressiveness and executability.
\citet{sergot1986british} demonstrated early on that legislation can be expressed, applied, and formalized in code, by encoding the British Nationality Act as a Prolog program.
Subsequent approaches introduced richer structure:
LegalRuleML~\cite{10.1145/2514601.2514603} provides an expressive XML standard for normative rules that satisfies legal domain requirements, designed to capture temporal dimensions, applies deontic operators, and allows linking legal rules to textual provisions.
The FLINT framework~\cite{breteler_flint_2023} models legal relations through actors, actions, preconditions, and normative effects.
\citet{vu_abstract_2022} represent legal text as Abstract Meaning Representation (AMR) graphs, and show that domain adaptation improves both AMR parsing and generation for legal documents.

\subsection{Law-to-Code Translation}
\paragraph{Traditional approaches.}
Early work on extracting decision models from legal text relied on manual encoding by domain experts, or semi-automated rule extraction using NLP techniques.
\citet{pertierra2017formalizing} explored automated parsing of tax code into logic representations, finding that semantic complexity can ``overwhelm'' existing natural language parsers' capabilities.
Prior work on automated extraction of DMN decision models from text, pre-dating LLMs, rely on different NLP methods, including coreference resolution, concept recognition and dependency parsing~\cite{etikala_text2dec_2020}, or deep learning methods~\cite{goossens_extracting_2023}.
These pre-LLM methods require task-specific architectures and typically substantial annotated training data.

\paragraph{LLM-based approaches.}
Recently, LLMs have been studied for legal formalization.
First, \citet{janatian2023fromtextto} use GPT-4 to generate structured representations from legislation for legal decision support, finding that 60\% of generated outputs being evaluated as equivalent or better than manually created ones.
More recently, \citet{zin2025machine} explore the use of LLMs for translating traffic rules into directly executable PROLOG code, studying in-context learning and fine-tuning.
However, this work relies on a pre-existing curated set of 20 traffic rules, substantial manual intervention for intermediate mapping to Logical English, and manually crafted PROLOG representations for evaluation.
Similarly, \citet{11190345} use in-context learning to generate executable Python representations of U.S. data breach notification laws, instantiated from a manually designed formal metamodel, and evaluated against a manually annotated benchmark dataset.
\citet{10.1007/978-3-031-46846-9_26} compare GPT-3 with a BERT-based approach for extracting Decision Requirements Diagrams (DRDs) for DMN, from short textual descriptions from government and university websites, and find that BERT outperforms GPT-3 overall, though GPT-3 shows promise when temperature is constrained.

Other recent work applies LLMs to various legal information extraction and knowledge representation tasks.
\citet{dalpont2025lost} extract deontic obligations from EU regulations using a theoretically grounded framework of obligation elements, finding strong performance on the AI Act but lower consistency across GDPR and DSA.
\citet{r2024legencomplexinformation} study generative approaches including BART, GPT-4o, and T5 for information extraction from legal sentences, finding that GPT's non-deterministic nature produces variable outputs.
\citet{nan2024combining} combine rule-based extraction with GPT-3.5 to extract legally required attributes from Dutch enforcement decisions, using "cheap" pattern matching to handle salient attributes, reducing the context sent to an LLM reserved for more difficult, less salient attributes.
\citet{billi2024fighting} use GPT-4o with few-shot learning to generate and revise Prolog rules for a legal expert system, proposing a human-in-the-loop approach that keeps the knowledge engineer in control of legal interpretation.
Finally, \citet{breton_leveraging_2025} and \citet{gray2024using} explore further applications, respectively extracting legal terms from traffic regulations and discovering legal factors from court opinions using LLMs.

\subsection{Semantic Role Labeling}
Semantic role labeling (SRL) annotates text with semantic roles (e.g., agents, goals, recipients), providing structured representations of events and their participants that answer questions such as \emph{``who does what to whom?''}
General-domain SRL corpora such as PropBank and FrameNet focus on broad language, and do not capture the complex syntax, domain-specific terminology, and normative structure of legal text.
\citet{humphreys-etal-2020-populating} demonstrate that semantic roles can bridge natural language and formal legal representations, using SRL to populate legal ontologies.
For Dutch legal text specifically, \citet{demaat2009nextstep} show 91\% accuracy on automated classification of provisions using syntactic parsing, and \citet{bakker-etal-2022-semantic} developed specialized SRL for Dutch law, comparing rule-based and transformer-based models for extracting semantic roles based on the FLINT framework.
In this paper, we use \textsc{FlintFiller-srl}~\citep{10.1145/3594536.3595124,bakker_semantic_2025}, a Dutch BERT model fine-tuned on 4,463 sentences of Dutch law labeled with four FLINT act roles: actors, actions, objects, and recipients.
We select this model because (i)~it is trained specifically on Dutch legal text, (ii)~its output categories align with our FLINT-based experimental design, and (iii)~pretrained models are publicly available.

\subsection{Evaluation and Code Generation}
The broader code generation literature shows that generation quality is sensitive to prompt design and example selection~\cite{austin2021programsynthesislargelanguage}, and that including signatures such as function name, input parameters, and output provides the most significant performance increase~\cite{11077752}. This directly motivates our use of \cond{}-specifications in prompts.
Evaluation approaches in this area combine semantic similarity metrics from NLP with outcome-based measures, common in tasks such as text-to-SQL generation~\cite{pinna2025redefining}.
Outcome-based evaluation measures functional equivalence by model execution, and requires well-defined, aligned inputs~\cite{zin2025machine}.
We adopt a dual strategy: graph-based similarity using graph kernels to quantify \emph{structural similarity} between generated and gold decision models, and \emph{outcome equivalence} to measure functional equivalence against pre-determined test scenarios.

\subsection{Positioning}
This work provides a controlled study of structured representations for generating executable decision models from legal text using LLMs.
Prior work has demonstrated feasibility of LLM-based legal formalization~\cite{janatian2023fromtextto,zin2025machine} and DMN extraction~\cite{etikala_text2dec_2020,goossens_extracting_2023}, but to our knowledge has not systematically compared the effects of source text enrichment against target constraints, nor examined in depth the relation between structural similarity and outcome equivalence, which reflects the recognized gap between syntactic form and semantic behavior in legal formalizations~\citep{sergot1986british}.

\section{Method}
\label{sec:method}

\subsection{Dataset}
\label{sec:dataset}
We make use of a unique real-world parallel corpus that aligns (a) legal text and (b) production-grade decision models that encode regulatory logic.
These models power the ``Environment and Planning Portal'' (\emph{Omgevingsloket}), a government platform where the public submits permit applications, checks requirements, and announces environmental activities governed by the ``Environment and Planning Act'' (\emph{Omgevingswet}), which regulates the physical environment of The Netherlands.
Unlike synthetic datasets, these are actively deployed models used daily by Dutch citizens and businesses for regulatory compliance.
The decision models are written in the Decision Model and Notation (DMN) standard in which logic is expressed as decision tables, whose rules map inputs to outputs.
Both decision models and legal text are shared publicly by the Government of the Netherlands.\footnote{\url{https://gitlab.com/koop/PR04/bruidsschat-teruglevering}}

The act covers \emph{environmental activities}, and the associated decision models cover two types:

\begin{itemize}
    \item \textbf{Outcome models} ($N=50$) determine under which regulatory outcome an environmental activity may fall, e.g.,
    is the activity exempt (\emph{niet van toepassing}),
    do general rules apply (\emph{algemene regels}),
    is there a notification obligation (\emph{informatieplicht}),
    or is a permit required (\emph{vergunningplicht})?
  \item \textbf{Requirements models} ($N=45$) determine whether submission requirements (SR) have been met for the latter two regulatory outcomes:
    \emph{SR Notification} ($N=23$) for activities that have a notification obligation, and
    \emph{SR Permit} ($N=22$) for activities that require a permit application.
\end{itemize}

We extract relevant legal articles from the \emph{Omgevingswet} by retrieving all articles that are explicitly linked to the decision models in their XML representations, and expand this set by following "internal cross-references" (\texttt{IntRef}-elements in the XML), adding additional articles that are referenced by the source article that belong to the same act.
This increases article coverage by 8.5\% (from 316 to 343 references).

\begin{table}[htbp]
\centering
\caption{Structural characteristics of gold-standard decision models (mean per model type).
Requirements models are larger on average than Outcome models.}
\label{tab:dmn-family-stats}
\small
\begin{tabular}{@{}llrrrr@{}}
\toprule
\textbf{Type} & \textbf{$N$} & \textbf{Nodes} & \textbf{Edges} & \textbf{Ext.Vars} & \textbf{Rules} \\
\midrule
Outcome & 50 & 23.0 \sd{39.6} & 36.0 \sd{65.9} & 9.5 \sd{15.8} & 58.4 \sd{103.2} \\
Requirements & 45 & 34.8 \sd{47.5} & 48.8 \sd{71.1} & 10.3 \sd{13.4} & 75.1 \sd{107.6} \\
\midrule
\textbf{Overall} & 95 & 28.6 \sd{43.9} & 42.0 \sd{68.7} & 9.9 \sd{14.7} & 66.3 \sd{105.6} \\
\bottomrule
\end{tabular}
\end{table}

To gain a better understanding of these decision models, we summarize their structural properties in Table~\ref{tab:dmn-family-stats}.
On average, Requirement models are larger than Outcome models (35 vs 23 nodes, with 75 vs 58 rules), with both having a similar numbers of input variables ($\sim$10).
The high standard deviations indicate substantial variations within each type.
Both types exhibit shallow rule logic: $\sim$90\% of rules contain a single condition, indicating chains of simple validations, rather than deeply nested Boolean compositions.

\subsection{Decision Models}
\label{sec:dmn-representation}
To facilitate LLM generation, we translate the verbose and lengthy original DMN format of the models to a simplified, lightweight JSON representation, using a deterministic mapping.
We keep nodes, edges, and all decision logic intact, and reduce the overall (textual) length of the models by a factor of 2.7$\times$ on average.
This is mostly XML syntax overhead, such as namespace declarations, verbose tag syntax, attribute formatting.
The resulting simplified decision models are directed acyclic graphs (DAGs) with three types of nodes:
(i) \textbf{external variables} representing inputs (e.g., ``is the building in a protected area?''),
(ii) \textbf{decision nodes}, the regulatory core of the models, which contain the decision tables (with \emph{``when X then Y''}-logic), using DMN hit policies that determine how rules are combined when multiple match (UNIQUE, FIRST, ANY, COLLECT),
(iii) a single \textbf{output node} that produces the final decision: categorical values for Outcome models, and boolean outputs for Requirements models (requirements are met or not).

\begin{figure}[t]
    \centering
    \includegraphics[width=\linewidth]{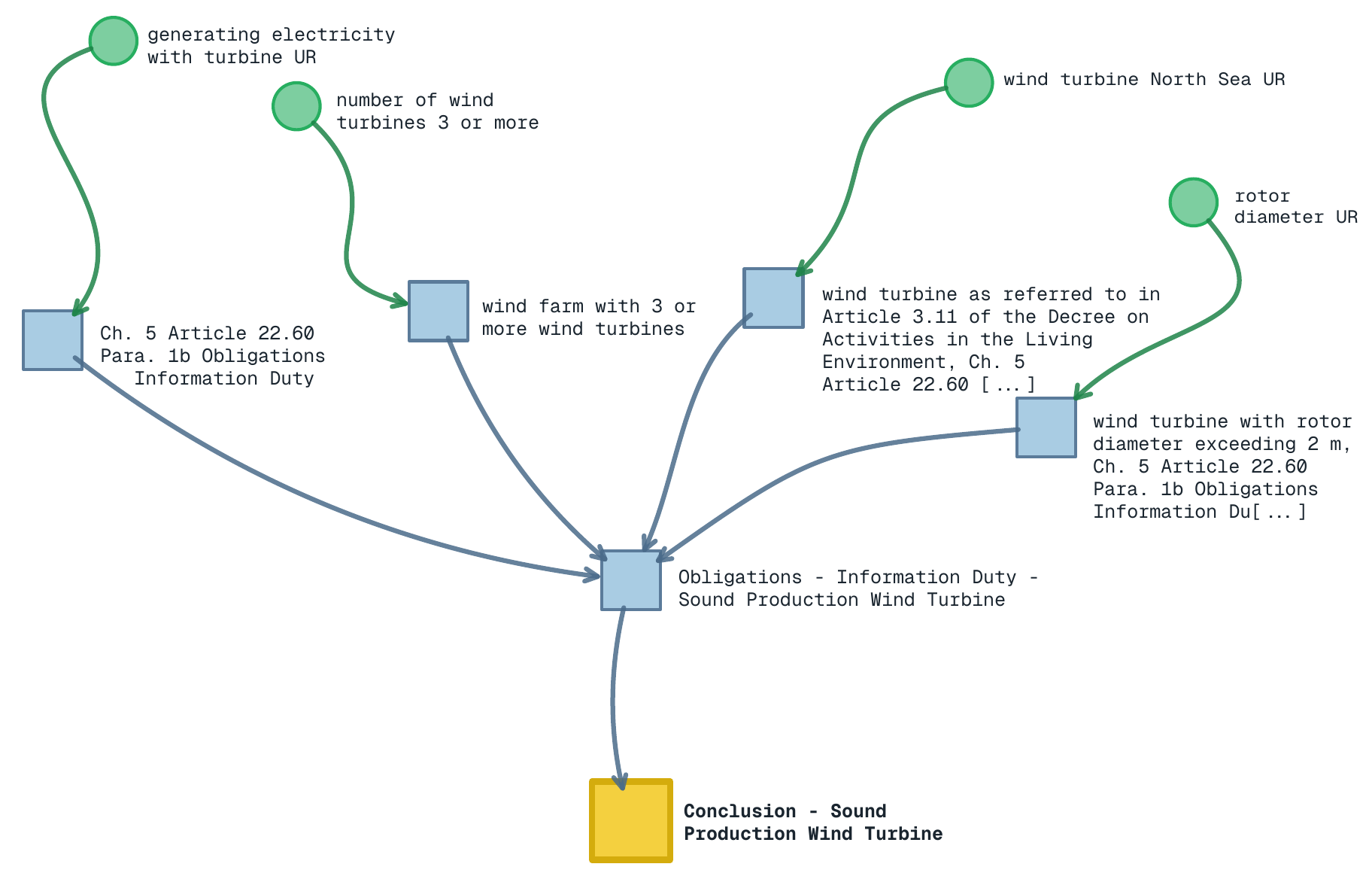}
    \Description{A directed acyclic graph of the GeluidProdWindturbine decision model. Four green input-node circles (whether the turbine generates electricity, belongs to a wind farm of three or more, is located in the North Sea, and has a rotor diameter over two metres) each connect through a blue pass-through decision-node square into a central blue aggregating decision node, which in turn feeds a single yellow output-node square producing the final regulatory outcome.}
    \caption{The \texttt{Outcome - GeluidProdWindturbine} decision model.
    Green circles represent input nodes (boolean questions), blue squares are decision nodes, and the yellow square produces the final regulatory outcome.}
    \label{fig:example}
\end{figure}

As an example, Figure~\ref{fig:example} shows a model determining whether a planned wind turbine triggers an \emph{information duty}, a requirement to notify authorities before construction, or whether only \emph{general rules apply}.
Four boolean inputs (green) describe the turbine:
whether it generates electricity,
belongs to a wind farm with $\geq$3 turbines,
is located in the North Sea, or
has a rotor diameter $>$2m.
Each input feeds a pass-through decision node (blue), which in turn feed the aggregating node.
The duty applies only when a specific combination of conditions holds; the turbine generates electricity, has a rotor exceeding 2m, is not part of a wind farm of three or more, and is not located in the North Sea, with any deviation short-circuiting the result to \texttt{false}.

\subsection{Input Representation}
\label{sec:augmentations}
We consider two input augmentations alongside the raw legal text: semantic role labels (\srl{}) extracted from legal text, and I/O specifications drawn from the target decision model.

\subsubsection{Semantic Role Labels (\srl{})}
\label{sec:srl-augmentation}
We enrich prompts by extracting semantic roles from the legal text, using flintfiller-\srl, a Dutch BERT model fine-tuned on a dataset of 4,463 Dutch law sentences annotated with four FLINT acts~\cite{10.1145/3594536.3595124}.
These FLINT acts describe:
\textbf{actors} (who performs the action? e.g., ``permit applicant'', ``municipality''),
\textbf{actions} (what is done? e.g., ``submit application'', ``construct building''),
\textbf{objects} (what is affected? e.g., ``building'', ``permit''), and
\textbf{recipients} (to whom/what, e.g., ``requester'').
We add flintfiller-\srl{}'s output to the prompt as additional metadata for the legal articles.
As an example, the activity \emph{KoelwaterLozen} (cooling water discharge) yields the following SRL labels from flintfiller-\srl{}:
  objects: \emph{koelwater} (cooling water), \emph{de maximale warmtevracht} (maximum heat load);
  actions: \emph{geloosd}, \emph{worden geloosd} (discharged);
  recipients: \emph{aan het college van burgemeester en wethouders} (to the municipal executive board).

\subsubsection{I/O specification (\cond)}
\label{sec:io-augmentation}
In addition, we enrich the LLM's input prompt by including \emph{input and output (I/O) specifications} from the decision models.
This mirrors findings from code generation, where function signatures (function name, input variables, outputs) provide stronger guidance than natural language elaboration alone~\cite{11077752}.
As shown in Figure~\ref{fig:example}, the decision models' inputs specify what input must be given to reach a decision, e.g., ``does the rotor diameter  exceed 2m?''
The output specifies the regulatory outcome or whether requirements have been met.
By providing I/O specifications, we fix the decision model's interface, while leaving the internal structure unconstrained; the LLM must still infer decision tables, rules, and intermediate nodes from the legal text.

\subsubsection{Combined representation (\srl{}+\cond)}
\label{sec:combined-augmentation}
The combined condition provides the labels from flintfiller-\srl{} (actors, actions, objects, recipients), and the decision model's I/O specification.
This allows us to test in how much enriching the source, or specifying the target help generation, and whether these approaches are complementary or redundant.

\subsection{Experimental Design}
\label{sec:protocol}

We evaluate the four input conditions described in Section~\ref{sec:augmentations}.
In each, the prompt includes an example decision model alongside one of the following:
\textbf{Text}: the target model's associated legal text only (baseline);
\textbf{Text+\srl{}}: legal text plus semantic role labels extracted by flintfiller-\srl{};
\textbf{Text+\cond{}}: legal text plus the target model's I/O specifications; and
\textbf{Text+\srl{}+\cond{}}: all of the preceding.

We apply 1-shot learning, i.e., we provide a single learning example in the LLM's prompt, which we select from the same decision model type (Outcome or Requirements), with additional quality filters (minimum 3 legal articles and at minimum 3 nodes).
For each of our 95 target decision models, we generate 5 independent runs per condition (changing only the randomly selected example) to assess generation stability.
For fair comparison across conditions, we fix the examples per run per model.
Our combination of conditions with independent runs yields an experimental size of 1,900 generations (95 models $\times$ 4 conditions $\times$ 5 runs).

We provide the LLM with a custom system message describing DMN decision-graph semantics, hit policies, and the required JSON output schema.
For each generation, we fill a user message template with an example decision graph, the associated legal article(s), and the target legal article(s), optionally enriched with \srl{} labels, \cond{} specifications, or both.
While the prompt fixes the output format and DMN semantics, it does not prescribe any domain-specific structure: the model must infer all variables, decision nodes, rule conditions, and dependencies from the legal text itself.

We use GPT-5.1, with temperature at 0.1 to balance determinism with structural variation, JSON mode response format to ensure valid output structure, and disabled chain-of-thought reasoning to prevent verbose intermediate steps that could interfere with structured generation.
We apply 1-shot learning rather than few-shot due to context length constraints: including multiple complete decision models with their associated legal articles would exceed practical token limits, and preliminary experiments showed diminishing returns beyond a single high-quality example.

We evaluate generated models along two dimensions:
\emph{structural evaluation} examines whether the generated model is similar to the gold model in structure, and
\emph{outcome equivalence} examines whether the generated model produces the same outputs as the gold, given same inputs.

\subsection{Structural Evaluation}
\label{sec:metrics-structural}
To evaluate how well the generated models match the gold models' structure, we use graph kernel similarity for quantitative comparison, and descriptive graph properties for interpretable analysis.

\subsubsection{Graph kernel similarity.}
We employ the Shortest-Path (SP) Kernel~\cite{borgwardt2005shortest}, which compares graphs based on the distribution of shortest paths between node pairs.
For decision graphs, this measures aspects such as depth and connectivity of decision chains from inputs to outputs.
We considered graph edit distance (GED), which turned out computationally intractable for some of the decision models in our dataset.
Finally, we also evaluated Graphlet kernels~\cite{shervashidze2009graphlet}, which compare graphs by counting small subgraph patterns (size 3-5 nodes), capturing local connectivity; however, these showed a clear ceiling effects (0.95+ similarity) that failed to discriminate between conditions, so we report only SP kernel results.
As node labels vary substantially across generations, we label nodes by their type, i.e., \emph{input}, \emph{decision}, or \emph{output}, enabling structural comparison across different node labels.

\subsubsection{Descriptive graph properties.}
We extract interpretable graph metrics, including size (number of nodes, edges), decision complexity (rules, conditions per rule), connectivity (in-degree, density), and depth (longest path, maximum width).
These enable direct and interpretable comparison of structural characteristics between gold and generated models.

\subsection{Outcome Evaluation}
\label{sec:metrics-behavioral}
Next to structural similarity, we aim to assess whether generated decision graphs are functionally equivalent, by evaluating their outcomes on controlled inputs.
We limit ourselves to the \cond{} and \srl{}+\cond{} conditions, as these have consistent inputs and outputs, enabling direct comparison.
We use a simplified graph executor that supports all decision table semantics from the DMN standard (FIRST, UNIQUE, ANY, COLLECT hit policies).

\subsubsection{Test inputs selection.}
\label{sec:test-input-sampling}
To select suitable decision models and their inputs, we select Outcome models with boolean inputs exclusively, at most 10, enabling $2^n$ exhaustive enumeration.
This yields a set of 24 of 50 Outcome models, with 2,712 test cases from exhaustively generating all input combinations.
For Requirements models, we extend our selection strategy to include string-typed input nodes, after discovering these are commonly simple categoricals.
Specifically, we found string variables typically fall in one of three categories:
(i) \textbf{categorical strings} checked via \texttt{contains(?, "value")}-logic in decision nodes (e.g., checks if a string contains \emph{``Start a new activity''} or \emph{``Change or expand\ldots''}),
(ii) \textbf{binned numeric strings} that are discretized in text (e.g., \emph{``Over 600 m$^3$''} or \emph{``Not over 600 m$^3$''}), and
(iii) \textbf{null-check strings} that test only for presence/absence of a value.
We extract all possible categorical values from string inputs via the \texttt{contains()}-patterns found in the (gold) decision rules.
This yields 34 out of 45 testable Requirements activities, with $\leq$1,024 combinations each, resulting in 10,368 Requirements cases.
In sum, we yield a set of 58 (out of 95) testable decision models across Requirements and Outcome models, with a total of 13,080 testable input variations.

\subsubsection{Metric}
\label{sec:equivalence-metric}
We compute outcome equivalence as the proportion of test inputs yielding identical decisions between gold and generated models.
For a single model $m$ with test set $X_m$:
\[
\text{Outcome}_m = \frac{|\{x \in X_m : G_{\text{gold}}(x) = G_{\text{gen}}(x)\}|}{|X_m|}
\]
We report \emph{macro-averaged} outcome equivalence, where we average the per-model equivalence rates, giving an equal weight to each decision model independent of their number of input variations.

\section{Results}
\label{sec:results}

\subsection{Structural Evaluation}
\label{sec:results-structural}

We measure structural properties of the generated models in two ways:
graph kernel similarity (\S\ref{sec:results-kernels}) for quantitative similarities, and
descriptive graph properties (\S\ref{sec:results-descriptive}) for interpretable and comparative analysis of structural properties between gold and generated models.

\subsubsection{Graph kernel similarity}
\label{sec:results-kernels}
Table~\ref{tab:main-results} presents graph kernel similarity between generated and gold decision models across the four experimental conditions.

\begin{table}[t]
\centering
\caption{Shortest-Path kernel similarity by condition and model type.
\cond{} significantly improves both types ($p<0.001$); \srl{} alone only helps Requirements ($p<0.001$).}
\label{tab:main-results}
\begin{tabular}{lcc}
\toprule
\textbf{Condition} & \textbf{Outcome} ($n=50$) & \textbf{Requirements} ($n=45$) \\
\midrule
Text            & 0.315 & 0.356 \\
Text+\srl        & 0.320 & 0.430 \\
Text+\cond       & \textbf{0.433} & 0.549 \\
Text+\srl+\cond   & 0.432 & \textbf{0.551} \\
\bottomrule
\end{tabular}
\end{table}

First, we note that Requirements models achieve consistently higher similarity across all conditions than Outcome models, suggesting they are easier to generate.
Semantic role labels alone (\srl) have different effects across the decision model types:
for Outcome models, the minor improvement is not statistically significant (0.320 vs 0.315, Wilcoxon $p=0.85$),
for Requirements models, however, \srl{} provides a significant improvement of +20.8\% (0.430 vs. 0.356, $p<0.001$).
I/O specifications (\cond) however, provide major improvements in structural similarity for both model types:
Outcome models improve by 37\% (0.433 vs. 0.315), and Requirements by 54\% (0.549 vs. 0.356), both $p<0.001$.
Adding \srl{} to \cond{} provides no additional benefit for either models with minor, non-significant differences to Text+\cond{} (Outcome: $p=0.71$; Requirements: $p=0.65$).

\subsubsection{Descriptive graph properties}
\label{sec:results-descriptive}

Next, we compare interpretable graph properties between generated and gold decision graphs.
Table~\ref{tab:comprehensive-stats} shows the graph statistics across three types: \emph{structure}, \emph{decision logic}, and \emph{depth}.

\begin{table*}[t]
\centering
\caption{Descriptive statistics comparing gold and generated graphs. Values are means across all activities within each type.}
\label{tab:comprehensive-stats}
\small
\begin{tabular*}{\textwidth}{@{\extracolsep{\fill}}llcccccccccc@{}}
\toprule
& & \multicolumn{5}{c}{\textbf{Outcome} ($n=50$)} & \multicolumn{5}{c}{\textbf{Requirements} ($n=45$)} \\
\cmidrule(lr){3-7} \cmidrule(lr){8-12}
\textbf{Category} & \textbf{Metric} & \textbf{Gold} & \textbf{Text} & \textbf{+\srl} & \textbf{+\cond} & \textbf{+\srl+\cond} & \textbf{Gold} & \textbf{Text} & \textbf{+\srl} & \textbf{+\cond} & \textbf{+\srl+\cond} \\
\midrule
Structure & Nodes & 23.0 & \textbf{8.1} & 8.1 & 5.9 & 5.9 & 34.8 & 10.8 & \textbf{12.2} & 11.8 & 11.5 \\
 & Edges & 36.0 & \textbf{22.4} & 21.6 & 12.5 & 12.9 & 48.8 & \textbf{25.1} & 24.6 & 20.9 & 21.0 \\
 & Input nodes & 9.5 & 14.3 & 13.8 & \textbf{9.3} & \textbf{9.3} & 10.3 & 17.3 & 15.2 & \textbf{10.3} & \textbf{10.3} \\
\midrule
Logic & Rules & 58.4 & 30.5 & \textbf{31.3} & 16.9 & 17.2 & 75.1 & 35.1 & \textbf{36.7} & 29.1 & 28.7 \\
 & Rules/node & 2.29 & 3.57 & 3.70 & \textbf{2.75} & 2.81 & 2.09 & 3.32 & 3.21 & \textbf{2.55} & 2.57 \\
 & Inputs/rule & 1.17 & 1.73 & 1.72 & \textbf{1.33} & 1.34 & 1.17 & 1.70 & 1.61 & \textbf{1.37} & \textbf{1.37} \\
\midrule
Depth & Depth & 4.1 & 2.9 & 3.0 & 3.2 & \textbf{3.2} & 4.7 & 2.7 & 2.9 & 3.6 & \textbf{3.6} \\
 & Max width & 11.9 & 14.4 & \textbf{13.9} & 9.8 & 9.8 & 14.8 & 17.4 & \textbf{15.4} & 10.8 & 10.8 \\
\bottomrule
\end{tabular*}
\end{table*}

In terms of \emph{structure}, we find that for (average) numbers of nodes, edges, and inputs, generated models are substantially smaller than their gold counterparts:
they contain only 26--35\% of gold nodes for Outcome (5.9--8.1 vs 23.0) and 31--35\% for Requirements (10.8--12.2 vs 34.8) models.
In contrast, Text and Text+\srl{} generate more input nodes than gold (around 1.5$\times$), despite producing fewer decision nodes overall.
For graphs generated with \cond-enriched prompts, with input and output nodes provided, we see as expected that generated Requirements models match gold input node counts exactly, with Outcome models showing a small difference (9.3 vs 9.5), due to the LLM failing to generate any input nodes for one model, despite being given its I/O specification.

Looking at \emph{decision logic} properties, we note that generated graphs show fewer rules in total, ranging from 16.9--31.3 for Outcome models (58.4 for gold), and between 28.7--36.7 for Requirements models (75.1 for gold), which can be attributed to their more compact structure with fewer nodes and edges, as we've seen above.

When it comes to rule complexity, however, we observe a different effect:
for conditions with no I/O specifications, models create on average more rules per node (between 3.2--3.7, vs 2.1--2.3 in gold), and also more inputs per rule (1.6--1.7 vs 1.2 gold).
Prompts with \cond{}-specifications produce simpler decision tables, more similar to those in the gold models (2.6--2.8 rules per node, and 1.3--1.4 inputs per rule).

In terms of \emph{depth}, all conditions produce shallower models than gold, at 2.9--3.2 for Outcome models (4.1 for gold); and 2.7--3.6 for Requirements models (4.7 for gold), indicating that generated models flatten deeper decision logic into fewer steps.

Interestingly, \cond-enriched prompts produce \emph{deeper} graphs than text only (3.2 vs 2.9 for Outcome; 3.6 vs 2.7 for Requirements), suggesting that the fixed inputs and outputs challenge the LLM to take a larger number of steps to connect inputs with output.
The text-only condition tends to yield wider graphs (ranging from 14.4--17.4), while \cond-enriched prompts produce narrower graphs (9.8--10.8): more similar to, but still less than gold models.

In sum, we find that overall, generated graphs tend to be smaller, shallower, and narrower than their gold counterparts.
In addition, including I/O specifications substantially changes the nature of generation:
representations without \cond{} generate wider graphs, with more inputs, and shallower depth.
Representations that include \cond{}, on the other hand, result in graphs with correct input counts (since they are fixed), deeper structure, but, still, fewer nodes overall.
While these smaller, shallower, and simpler models may suggest inaccurate model generation, we turn to outcome equivalence testing for functional validation in Section~\ref{sec:results-outcome-equivalence}.

\subsubsection{Identity nodes in gold models}
\label{sec:identity-nodes}
The gap in node counts between gold and generated models discussed above might suggest incomplete generation.
However, upon inspection we find that a large number of nodes in gold models are so-called \emph{identity nodes}: decision tables that simply pass through their input unchanged.

We identify identity nodes by checking whether a node's decision logic consists solely of rules that map each input value to itself.
Across all gold models, we find that 54.7\% of Outcome nodes and 45\% of Requirements nodes are identity nodes.
To illustrate, we find that gold models follow a consistent naming convention
that encodes processing stages: User Requirement (UR) nodes capture user input
(e.g., \texttt{X UR}), Case Requirement (CR) nodes perform null-check validation
(e.g., \texttt{X beantwoord CR} with \texttt{not(null)} logic), and pass-through
nodes apply identity transformations (e.g., \texttt{X}, \texttt{X herbruikbare set}).
A typical chain from input to output thus contains 3--4 nodes, of which only
one (the CR node) contains actual logic: checking whether the user provided a value.

We find that generated models reduce these identity nodes by 83--100\%, while preserving the essential null-checking logic.
Rather than encoding \texttt{not(null)} checks in separate CR nodes followed by identity chains, generated models typically incorporate null-checking directly into decision rules (e.g., \texttt{if X = null then false else ...}), yielding functionally equivalent but structurally more compact models.

These identity nodes in gold models reflect a \emph{modeling convention} rather than semantic necessity.
The UR/CR naming pattern and pass-through chains encode workflow stages (from user input to validation to processing) that could be expressed differently.
Generation of models implicitly discovered this redundancy: and as we will show in Section~\ref{sec:results-outcome-equivalence}, 33\% of generated models achieve full outcome equivalence with gold models on at least one of five runs, and 50\% reach $\geq$90\% equivalence, despite having 65–74\% fewer nodes.
Here, smaller generated graphs reflect more efficient generation, not incomplete generation.

\subsection{Outcome Equivalence}
\label{sec:results-outcome-equivalence}
Next to structural similarity, we evaluate whether generated decision models produce \emph{equivalent outputs} when executed on the same inputs.
This tests functional correctness: do the generated decision models reach the same conclusion as the gold models?

\subsubsection{Methodology}
We apply the methodology described in \S\ref{sec:test-input-sampling} to obtain 58 testable decision models (24 Outcome, 34 Requirements) and 13,080 test cases.
We limit outcome equivalence testing to generated models with \cond-enriched prompts (i.e., Text+\cond{} and Text+\srl+\cond{} conditions), as these have fixed inputs that allow direct comparison to gold models with identical inputs.
We execute both gold and generated models on all possible input combinations across all 5 runs.

For Outcome models with string-based outputs, we apply a simple keyword-based classification method to categorize diverse outputs into four classes that represent the different regulatory outcomes:
\emph{NotApplicable} indicates an activity does not apply,
\emph{PermitRequired} indicates a permit is required,
\emph{GeneralRulesApply} indicates that general rules apply,
\emph{NotificationRequired} indicates an obligation to notify before the start of the activity.

\subsubsection{Results}
Table~\ref{tab:outcome-agreement} summarizes outcome equivalence across both model types, reporting macro-averaged agreement as described in \S\ref{sec:equivalence-metric}.

\begin{table}[t]
\centering
\caption{Outcome equivalence between gold and generated decision models by model type. Values are macro-averaged (mean agreement per decision model).}
\label{tab:outcome-agreement}
\begin{tabular}{llcc}
\toprule
\textbf{Type} & \textbf{$n$} & \textbf{Text+\cond} & \textbf{Text+\srl+\cond} \\
\midrule
Outcome & 24 & 40.2\% & 42.6\% \\
Requirements & 34 & 59.2\% & 60.4\% \\
\midrule
\textbf{Combined} & 58 & 51.3\% & 53.1\% \\
\bottomrule
\end{tabular}
\end{table}

We find that Requirements models achieve substantially higher outcome equivalence (ranging from 59.2\% to 60.4\%) than Outcome models (40.2\%--42.6\%).
Text+\srl+\cond{} achieves higher agreement than Text+\cond{} (+2\%), suggesting semantic role labels provide a modest boost when I/O specifications are already provided.

After further inspection, we find that the majority of the Outcome models in our set (17/24) contain a placeholder input variable named \emph{``vaste waarde FALSE''} (fixed value FALSE) that always evaluates to FALSE, implementing constant-output patterns rather than actual decision logic.
Filtering these placeholder models from the Outcome results yields higher agreement at 57.3\% (for Text+\srl+\cond{}).

Macro-averaged agreement masks substantial variation across runs.
Examining best-run performance per model (i.e., the highest agreement among the 5 runs with different examples),
33\% of generated models achieve full outcome equivalence (19/58) and 50\% reach $\geq$90\% equivalence (29/58) under Text+\srl+\cond{}.
This suggests the LLM is capable of producing functionally correct models for roughly half of our targets, but does so inconsistently: example selection determines whether a given run succeeds.

\subsubsection{Discussion}
Our results should be read in context.
The evaluation is conservative: rare input combinations receive equal weight as common combinations, and exhaustive sampling ignores real-world input correlations.
The gold standard is demanding: production-grade decision models that carry legal authority, actively deployed in a government portal, and the product of careful expert modeling.
51–53\% macro-averaged agreement, with best-run performance of 33\% full equivalence and 50\% $\geq$90\%, shows the task is tractable, but legal stakes mean human review remains essential.

Beyond these caveats, comparing generated outputs to gold standards requires careful handling of surface form variations.
We found that \srl-enriched prompts produce different outputs: boolean-like conditions were sometimes generated as Dutch strings (\emph{``Ja''}/\emph{``Nee''}), and more verbose textual conclusions and explanations were typically generated.
Our evaluation normalizes these by mapping these Dutch strings to booleans, and applying keyword-based pattern matching for output classification.

Finally, we observe that outcome equivalence and structural similarity are complementary:
of the generated models with high structural similarity to gold models (SP $\geq$ 0.5), 36\% have low outcome equivalence (<50\% agreement).
Conversely, 39\% of generated models with high outcome equivalence ($\geq$50\%) have low structural similarity (SP $\leq$ 0.5).
We illustrate this complementarity with examples in the following section.

\subsection{Qualitative Examples}
\label{sec:results-qualitative}

To complement our quantitative findings, we present two illustrative examples that reveal the complementary properties of structural similarity and outcome equivalence, mentioned in Sections~\ref{sec:results-structural} and \ref{sec:results-outcome-equivalence} above.
Both are Requirements models, evaluated under the Text+\srl+\cond{} condition.

\subsubsection{GeluidProdWindturbine: high structure, low outcome}
\label{sec:example-geluid}

\begin{figure}[t]
    \centering
    \begin{subfigure}[b]{0.9\linewidth}
        \centering
        \includegraphics[width=\linewidth]{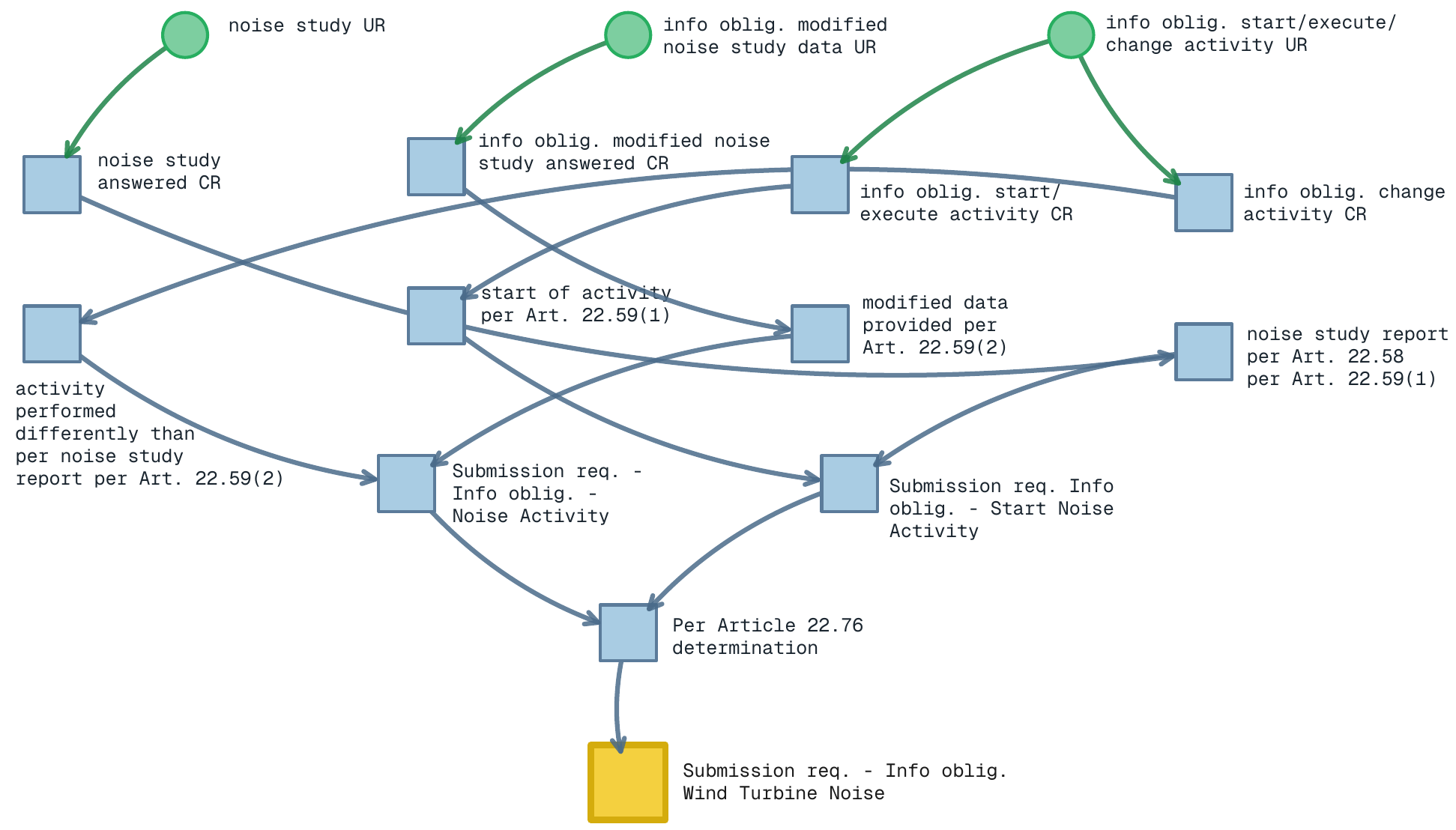}
        \Description{Gold-standard decision graph for the GeluidProdWindturbine requirements model, containing eleven decision nodes arranged in several parallel chains from input nodes to a single output node.}
        \caption{Gold standard (11 decision nodes)}
    \end{subfigure}
    \begin{subfigure}[b]{0.9\linewidth}
        \centering
        \includegraphics[width=\linewidth]{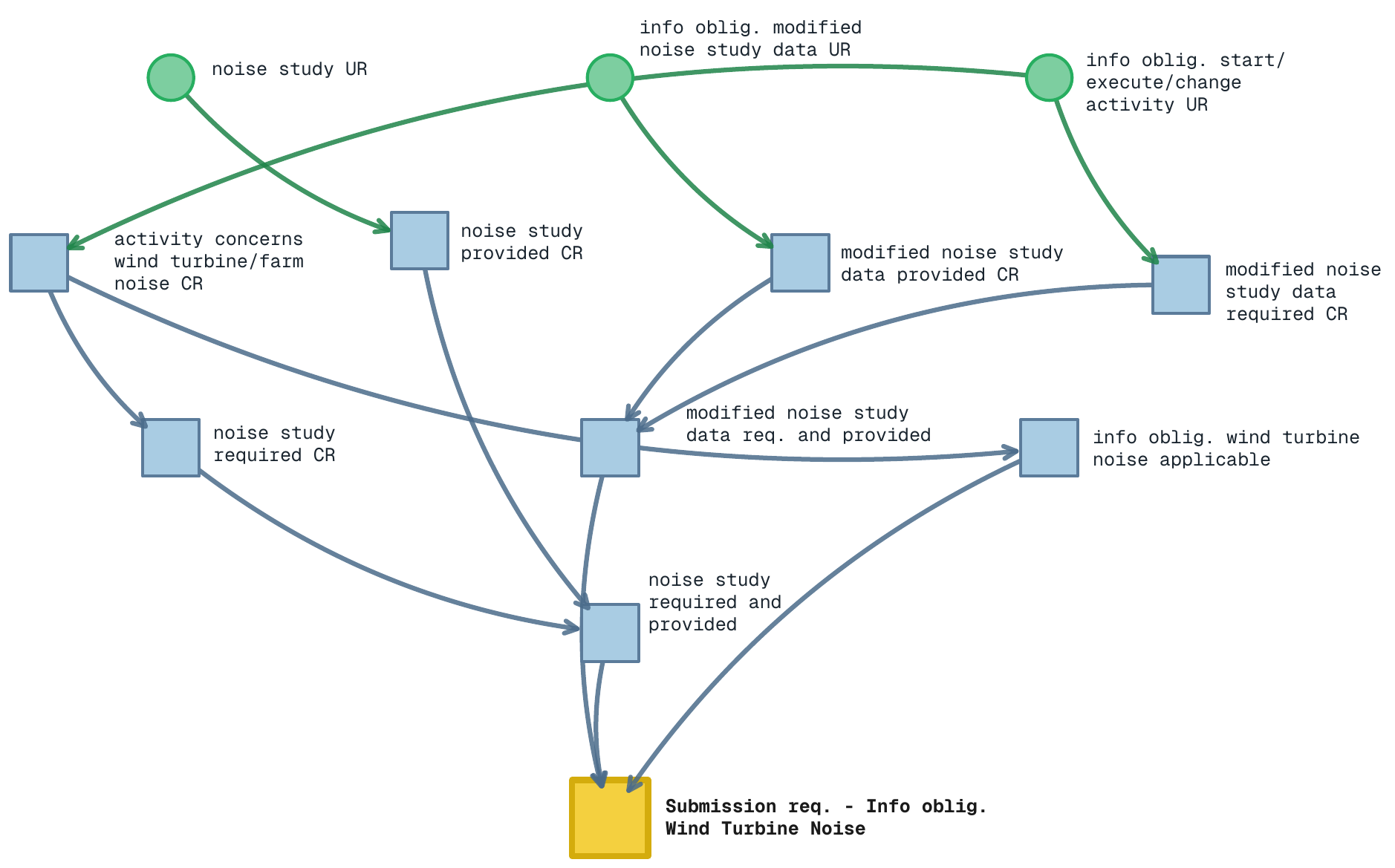}
        \Description{Generated decision graph for the same model, containing eight decision nodes in a visually similar parallel arrangement to the gold standard, but with different rule content inside the nodes.}
        \caption{Generated (8 decision nodes)}
    \end{subfigure}
    \caption{GeluidProdWindturbine: High structural similarity (SP=0.85 with 8 nodes) to the gold standard (11 nodes) but low outcome equivalence: 30\%.
    The generated graph captures the overall structure but encodes different decision rules.}
    \label{fig:example-geluid}
\end{figure}

The GeluidProdWindturbine (wind turbine noise) requirements model determines whether the submission requirements for the information obligation regarding a wind turbine's noise production have been met.
Figure~\ref{fig:example-geluid} shows that the generated model achieves high structural similarity (SP=0.85), showing visually comparable structure to the gold standard.
However, outcome equivalence is only 30\%: although the structure matches and inputs and outputs are identical, the generated rules encode different logic.

This demonstrates that structural similarity is not sufficient for outcome equivalence: the model successfully infers decision structure from legal text, but fails to correctly translate the logic within the decision nodes.

\subsubsection{AlarminstallatieHebben: removing redundant inputs}
\label{sec:example-alarm}

\begin{figure}[t]
    \centering
    \begin{subfigure}[b]{0.85\linewidth}
        \centering
        \includegraphics[width=\linewidth]{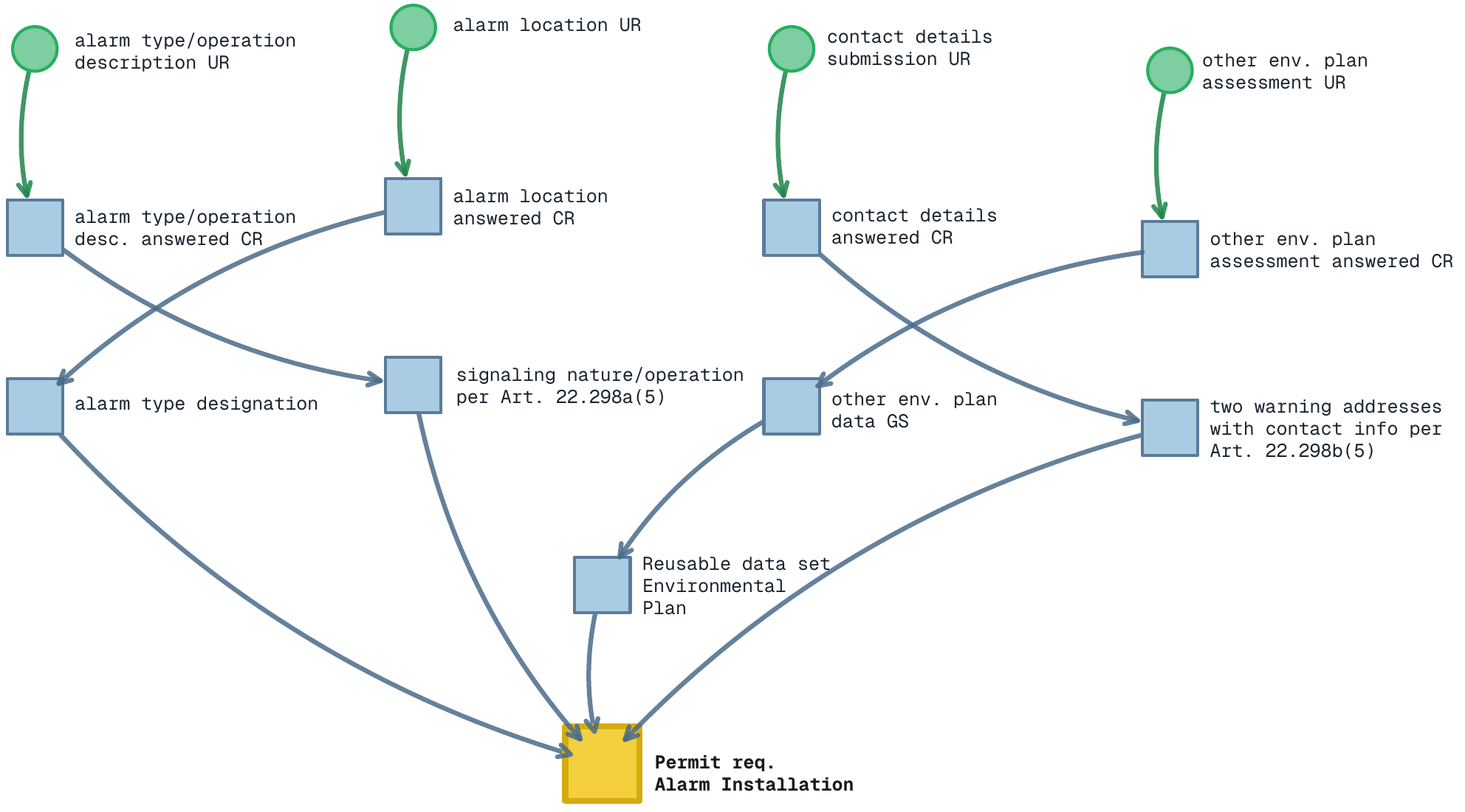}
        \Description{Gold-standard decision graph for the AlarminstallatieHebben requirements model, containing nine decision nodes. Four input nodes each flow through a UR node and then a CR node performing a not-null check before reaching the downstream decision logic, producing a characteristic UR/CR pass-through chain structure.}
        \caption{Gold standard (9 decision nodes)}
    \end{subfigure}
    \begin{subfigure}[b]{0.85\linewidth}
        \centering
        \includegraphics[width=\linewidth]{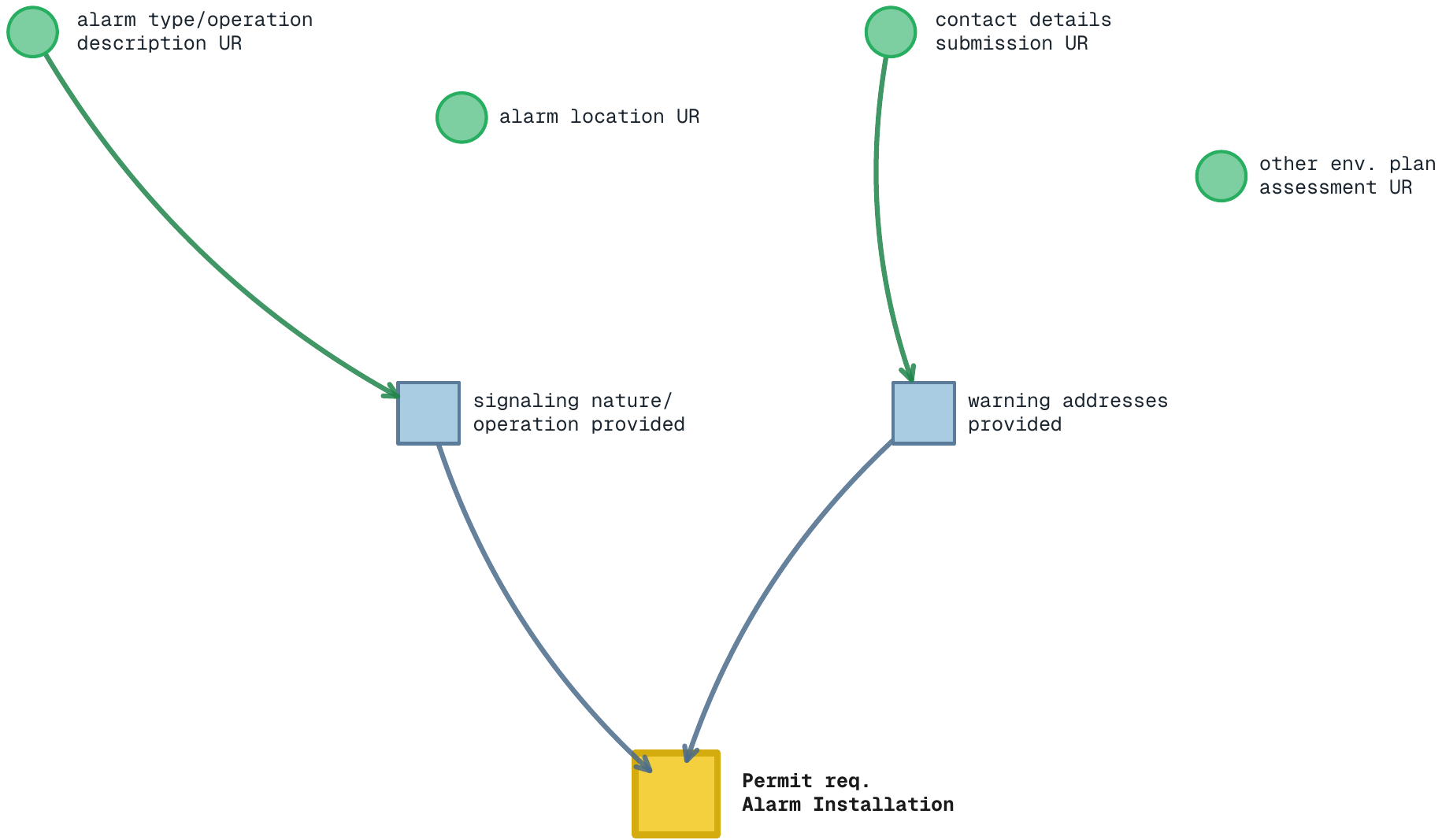}
        \Description{Generated decision graph for the same model, containing only two decision nodes. Two of the four specified input nodes are rendered but disconnected from any decision node; the remaining two input nodes each connect directly to a single decision node, eliminating the UR/CR pass-through chains present in the gold standard.}
        \caption{Generated (2 decision nodes)}
    \end{subfigure}
    \caption{AlarminstallatieHebben: Low structural similarity (0.37) but perfect outcome equivalence (100\%).
    The generated model correctly identifies and removes redundant boolean inputs, simplifying from 9 to 2 decision nodes.}
    \label{fig:example-alarm}
\end{figure}

Finally, the AlarminstallatieHebben (alarm installation) model determines whether the permit requirements for an alarm system have been met.
While Figure~\ref{fig:example-alarm}b shows low structural similarity (SP=0.37), the generated model achieves \emph{perfect} outcome equivalence (100\%).
Notably, the generated graph only has 2 decision logic nodes, compared to 9 in the gold model.
Two of four of the specified inputs: \emph{alarm location UR} and \emph{other env. plan assessment UR} (from \cond{}) are rendered, but disconnected from any (decision) nodes.

These same nodes in the gold model (Figure~\ref{fig:example-alarm}a) illustrate the UR/CR chain described in Section~\ref{sec:identity-nodes}: each of the four inputs flows through a ``UR'' node (user requirement), then a ``CR'' node (case requirement performing a \texttt{not(null)} check), before reaching decision logic.
The generated model (Figure~\ref{fig:example-alarm}b) eliminates this overhead entirely, connecting only the two remaining input nodes to two distinct decision nodes.
It correctly identifies that for two boolean inputs, both \texttt{true} and \texttt{false} pass the \texttt{not(null)}-check, meaning these inputs are effectively constant.

This example illustrates a \emph{valid simplification} by the LLM, not an error.
This confirms that low structural similarity does not necessarily indicate incorrectness, and shows that human-crafted gold models may contain redundant complexity that automated generation effectively eliminate.

\subsection{Legal Text Complexity Effects}
\label{sec:text-complexity}
To deepen our quantitative analysis, we examine whether properties of the source legal text predict model generation quality.

\begin{table}[t]
\small
\centering
\caption{Graph kernel similarity (SP) by text complexity features (tertile splits).
Complex legal texts substantially reduce Outcome quality, while Requirements models are largely immune to cross-references and recital length.}
\label{tab:text-complexity}
\begin{tabular}{lllcc}
\toprule
\textbf{Feature} & \textbf{Level} & \textbf{Range} & \textbf{Outc.} & \textbf{Reqs} \\
\midrule
Avg. sentence len & Low    & 10--14 words & 0.56 & 0.59 \\
                & Medium & 14--17 words & 0.48 & 0.58 \\
                & High   & 17--24 words & 0.22 & 0.39 \\
\midrule
Recital length & Low    & 0--300 words   & 0.64 & 0.52 \\
               & Medium & 300--600 words & 0.35 & 0.62 \\
               & High   & 600+ words     & 0.38 & 0.52 \\
\midrule
Cross-references & Low    & 0--4 refs  & 0.60 & 0.53 \\
                 & Medium & 5--8 refs  & 0.39 & 0.58 \\
                 & High   & 9+ refs    & 0.37 & 0.54 \\
\midrule
List items & Low    & 0--3 items   & 0.44 & 0.64 \\
           & Medium & 4--14 items  & 0.52 & 0.48 \\
           & High   & 16+ items    & 0.39 & 0.57 \\
\bottomrule
\end{tabular}
\end{table}

Table~\ref{tab:text-complexity} shows graph kernel similarity stratified by text features that represent:
\textbf{length}: \emph{(average) sentence lengths} of the legal text, and of the accompanying \emph{recital text}, as a proxy to the complexity of the rules expressed in the legal text,
\textbf{external dependencies}: the number of \emph{cross-references} to other laws in text,
and \textbf{structure}: the number of list items per legal article, hypothesizing that structured lists map more directly to discrete decision logic.

We observe different patterns between model types:
First, average sentence length is negatively correlated with structural similarity for both model types: longer sentence lengths reduce structural similarity by 61\% for Outcome models (0.56 to 0.22), and by 34\% for Requirements models (0.59 to 0.39).
Recital text lengths and number of cross-references to other laws only affect Outcome models.

Articles with a higher number of cross-references hurt Outcome models and reduce similarity on average by 39\%, but have practically no effect on Requirements (0.53 to 0.54).
Similarly, long recitals, which may indicate complex regulation, reduce Outcome similarity by 41\% but show no comparable effect for Requirements.

Structural features show a more nuanced pattern: list items correlate negatively with generation quality overall ($\rho = -0.23$), but Outcome models peak at a medium number of items, while Requirements models show a U-shape, with higher similarities at either few or many items.

\subsection{The Role of Example Selection}
\label{sec:results-stability}
Finally, with 5 independent runs per target, each using a different randomly-selected gold example, but held fixed per run across the conditions, we assess how example selection affects generation quality.
We examine:

\begin{itemize}
  \item \textbf{Example similarity (\texttt{ExSim})}: SP kernel similarity between the input example and the gold target; higher values indicate an "easier" task.
    \item \textbf{Generation similarity (\texttt{GenSim})}: SP kernel similarity between the generated model and the gold target; higher values indicate generations closer to the target.
    \item \textbf{Example consistency (\texttt{ExCon})}: mean pairwise SP kernel similarity among the 5 examples for a target across runs; higher values indicate more homogeneous example sets.
    \item \textbf{Generation variance (\texttt{GenVar})}: standard deviation of \texttt{Gen\-Sim} across the 5 runs per target. Higher values indicate less consistent generation quality.
\end{itemize}

Table~\ref{tab:example-effects} shows two correlations: between \texttt{ExSim} and \texttt{GenSim}, to study whether examples similar to the target yield better generations, and between \texttt{ExCon} and \texttt{GenVar}, to study whether consistent example sets produce consistent generations.

\begin{table}[t]
    \centering
    \caption{Effect of examples on generation (Spearman $\rho$).
    I/O specifications amplify the benefit of good examples, and reduce sensitivity to example variation.}
    \label{tab:example-effects}
    \begin{tabular}{lcccc}
    \toprule
    & \multicolumn{2}{c}{\texttt{ExSim}$\rightarrow$\texttt{GenSim}} & \multicolumn{2}{c}{\texttt{ExCon}$\rightarrow$\texttt{GenVar}} \\
    \cmidrule(lr){2-3} \cmidrule(lr){4-5}
    \textbf{Condition} & $\rho$ & $p$ & $\rho$ & $p$ \\
    \midrule
    Text & +0.12 & $<$0.05 & +0.50 & $<$0.001 \\
    Text+\srl & +0.20 & $<$0.001 & +0.28 & $<$0.01 \\
    Text+\cond & +0.36 & $<$0.001 & +0.14 & 0.19 \\
    Text+\srl+\cond & +0.38 & $<$0.001 & +0.12 & 0.26 \\
    \bottomrule
    \end{tabular}
\end{table}

First, we note that in the conditions without I/O constraints, example-to-target similarity (\texttt{ExSim}) only weakly predicts generation quality ($\rho=0.12/0.20$), while similarity among the examples in the set across runs (\texttt{ExCon}) strongly predicts generation variance ($\rho=0.50/0.28$).
Counterintuitively, more homogeneous example sets correlate with more varied generations, suggesting that similar examples may reinforce patterns that diverge from the target.
 
With I/O specifications, this pattern changes: example similarity now more strongly predicts generation quality ($\rho=0.36/0.38$), while variance becomes more independent from example selection ($\rho=0.14/0.12$, $p>0.19/0.26$).
The latter is intuitive: fixed I/O leaves less room for example variation to affect the output.
But the former, example similarity more strongly predicting generation quality, is harder to explain mechanically, and suggests I/O specifications help the model make better use of examples for the parts of the graph that remain unconstrained.

\section{Conclusion}
\label{sec:conclusion}
We evaluated LLM-based generation of executable decision models from legal text, comparing four representations across 95 real-world decision models that power the Omgevingsloket government portal.
Our key finding is that I/O specifications provide the dominant improvement in generation quality.
Text+\cond{} achieves +37\% and +54\% relative improvements over the text-only baseline.
Semantic role labels (\srl) alone provide minimal benefit for Outcome models, and modest improvements for Requirements models.

Outcome equivalence testing shows generated models achieve 51–53\% average agreement with gold decision models, with 50\% reaching $\geq$90\% equivalence under best-run conditions.
Generated models are consistently smaller and simpler, highlighting how LLMs can successfully identify redundancy and reduce complexity of hand-crafted decision models.
Text complexity analysis reveals that longer sentences and more cross-references predict lower generation quality, more so for Outcome than Requirements models.

Beyond improving generation quality, I/O specifications increase the benefit of good examples, and reduce sensitivity to example variation, suggesting LLMs benefit more from knowing \emph{what} to generate than from additional contextual information~\cite{11077752}.

A key contribution is our evaluation on real-world data: our 95 decision models are actively deployed in a government portal used by Dutch citizens and businesses for regulatory compliance.
This real-world setting reveals practical aspects that synthetic or academic datasets may miss, such as the prevalence of workflow conventions (e.g., UR/CR-chains).

\subsection{Limitations}
We acknowledge several limitations to our experimental design.
First, our scope is limited to 95 decision models from a single jurisdiction, language, and domain, which may affect generalizability.
However, our dataset provided a unique opportunity for systematically analyzing LLM-based decision model generation in a real-world setting.
In addition, our methodology is designed to be in principle language-, domain-, and jurisdiction-agnostic.

We test a single LLM (GPT-5.1) with 1-shot prompting, partly due to cost and context constraints, but also because a single (model, prompting) setup isolates the effect of input representations without confounding it with model- or prompting-level variation.
Extending to multiple models or more elaborate prompting setups (e.g., chain-of-thought or multi-agent) is a natural direction for follow-up research that our data and code release supports.

Our scope of structural evaluation is likewise limited: kernel-based similarity metrics may miss nuanced semantic equivalences between models.
We attempted other metrics, including GED (which turned out to be intractable for certain graphs) and Graphlet kernels (which showed a plateau, and lacked distinguishing capability).

Finally, the \cond{}-condition requires I/O specifications in advance, which may seem contrived.
However, we argue that specifying I/O upfront is less contrived than it may appear: identifying relevant inputs (what does a citizen need to provide?) and possible regulatory outcomes (permit / notification obligation / exemption), is the kind of analysis a domain expert would do anyway, and is analogous to specifying a function signature, before implementing it in code~\cite{11077752}.
Our results show that this effort upfront may be a worthwhile precondition rather than an obstacle.

\subsection{Implications}
\label{sec:implications}
Our findings show that I/O specifications are instrumental in reliably generating decision models, while semantic enrichments provide less benefit, in line with similar findings in LLM-based code generation~\cite{11077752}.
This has practical implications, such as prioritizing interface definitions over elaborate parsing of legal text.
This also points towards a specific human-LLM division of labor: I/O specifications draw on legal domain expertise (which inputs are relevant, which outcomes are possible), while generating decision logic draws on technical modeling expertise (decision tables, hit policies, graph structure).
Identifying legally relevant features of a wind turbine application, and authoring executable, production-grade DMN requires distinct expertise and skills, rarely held by the same person.
If LLMs can bridge this modeling gap, the bottleneck shifts from modeling expertise to domain knowledge.

At the same time, the 51–53\% outcome equivalence shows LLM output cannot replace manually crafted models, and for models carrying legal authority, human review remains necessary irrespective of these numbers.
Here too, our findings suggest a human-assisted workflow rather than autonomous generation, where domain experts define I/O specifications, LLMs generate initial decision models from legal text, and experts then verify and refine them.

Practitioners can then expect structurally smaller models that often remain outcome-equivalent: our analysis shows 45–55\% of gold-model nodes are pass-through (UR/CR-chains) that do not affect the output, which we characterize as convention rather than semantic necessity.
LLMs do not reproduce these chains.
Whether this simplification is desirable depends on what purpose this structure serves, beyond the computed output.

Finally, the complementary nature of structural and outcome similarity affects validation: structural similarity alone cannot guarantee legal accuracy, which suggests real systems must implement outcome-based testing alongside structural similarity metrics.
Our error analysis also informs validation strategy:
without I/O constraints, models infer more input nodes and over-generate parallel structure.
With I/O specifications, models produce compact structures with correct interfaces.
Unconstrained outputs require checking whether inputs match requirements; constrained outputs require verifying that all necessary decision logic is captured.
Different input representations thus call for different review strategies.

\begin{acks}
David Graus is partly funded by \grantsponsor{ICAI}{ICAI (AI for Open Government Lab)}{https://icai.ai/}.
Views expressed in this paper are not necessarily shared or
endorsed by those funding the research.
The author thanks Anne Schuth and Damiaan Reijnaers for helpful discussions, and the anonymous reviewers for their constructive feedback.
Claude (Anthropic) was used to assist with editing, and with scripts for data processing and analysis.
All content remains the author's sole responsibility.
\end{acks}

\bibliography{bibliography}

\end{document}